%% file: l4dc2024.tex
\documentclass[11pt]{article}
\usepackage{authblk}
\usepackage{times}
\usepackage{amsmath,amssymb,amsfonts}
\usepackage{xcolor}
\usepackage{hyperref}
\usepackage{verbatim}   
\usepackage{algorithm,algorithmic}
\usepackage{mathtools}
\usepackage{siunitx}
\usepackage{pgfplots}
\usepackage{pgfplotstable}
\usepackage{multirow}
\usepgfplotslibrary{fillbetween}
\usepackage{booktabs}
\usepackage{wrapfig}
\usepackage[numbers,square,sort&compress]{natbib}
\pgfplotsset{compat = newest}
\usetikzlibrary{arrows,positioning,shapes,intersections,external,patterns,calc,fit,decorations,decorations.markings} 
\newtheorem{defn}{Definition}
\newtheorem{rem}{Remark}

\newtheorem{assum}{Assumption}

\newcommand\tran{\mkern-2mu\raise1.25ex\hbox{$\scriptscriptstyle\top\hspace{0.5mm}$}\mkern-3.5mu}
\newcommand{\R}{\mathbb{R}}

\newcommand{\C}{\mathcal{C}}

\newcommand{\X}{\mathcal{X}}

\newcommand{\Z}{\mathcal{Z}}
\newcommand{\bm}[1]{{\boldsymbol{#1}}}

\newcommand{\z}{\bm z}

\newcommand{\x}{\bm x}
\newcommand{\dxdt}{\frac{\partial \x}{\partial t}}

\renewcommand{\u}{\bm{u}}
\newcommand{\y}{\bm{y}}

\usepackage[noabbrev]{cleveref} 
\crefname{rem}{Remark}{Remarks}
\crefname{exam}{Example}{Examples}
\crefname{assum}{Assumption}{Assumptions}
\crefname{prop}{Proposition}{Propositions}
\crefname{propy}{Property}{Properties}
\crefname{cor}{Corollary}{Corollaries}
\crefname{lem}{Lemma}{Lemmas}
\crefname{section}{Section}{Sections}
\crefname{thm}{Theorem}{Theorems}
\crefname{alg}{Algorithm}{Algorithms}
\crefname{defn}{Definition}{Definitions}
\crefname{figure}{Fig.}{Fig.}
\Crefname{figure}{Figure}{Figures}
\crefname{equation}{}{}

\begin{document}
\title{Physics-Constrained Learning for PDE Systems \\ with Uncertainty Quantified Port-Hamiltonian Models}
% \author{Kaiyuan Tan}
% \author{Peilun Li}
\author{Kaiyuan Tan, Peilun Li, Thomas Beckers\thanks{The authors are with the Department of  Computer Science, Vanderbilt University, Nashville, TN 37212, USA. Email:{\tt\small \{kaiyuan.tan, peilun.li, thomas.beckers\}@wustl.edu}}
}
\date{}
\maketitle

\begin{abstract}%

Modeling the dynamics of flexible objects has become an emerging topic in the community as these objects become more present in many applications, e.g., soft robotics. Due to the properties of flexible materials, the movements of soft objects are often highly nonlinear and, thus, complex to predict. Data-driven approaches seem promising for modeling those complex dynamics but often neglect basic physical principles, which consequently makes them untrustworthy and limits generalization. To address this problem, we propose a physics-constrained learning method that combines powerful learning tools and reliable physical models. Our method leverages the data collected from observations by sending them into a Gaussian process that is physically constrained by a distributed Port-Hamiltonian model. Based on the Bayesian nature of the Gaussian process, we not only learn the dynamics of the system, but also enable uncertainty quantification. Furthermore, the proposed approach preserves the compositional nature of Port-Hamiltonian systems.
\end{abstract}

% \begin{keywords}%
%  physics-constrained learning, scientific machine learning, Gaussian processes
% \end{keywords}

\section{Introduction}

Modeling the dynamics of partial differential equation (PDE) systems plays a pivotal role in making safe and efficient control decisions, as they can forecast future states of the desired systems. The applicability of such models extends across various domains, including but not limited to flexible robots \citep{zheng2022pde}, heat transfer processes \citep{zobeiry2021physics}, and electrical systems \citep{cieslik2021mathematical}. However, accurately modeling nonlinear PDE systems remains challenging, particularly due to their inherent complexity. A salient aspect of this complexity is the intricate modeling of nonlinearities within these systems, a task that is challenging and critical for accurate models. Traditional first-principle modeling methods often struggle with highly nonlinear effects, leading to solutions that may not fully capture the underlying physical reality. To address the lack of knowledge of complex systems, data-driven methods have been proposed to address state forecasting tasks, e.g., \cite{long2018pde,long2019pde,bar2019learning,wu2020data,stephany2022pde}.
However, those pure data-driven learning approaches pose new questions regarding the efficiency, safety, and physical correctness of learned models~\citep{hou2013model}. Due to the neglect of physical priors, the pure-learning approach suffers from many disadvantages, e.g., limited trustworthiness  \citep{feng2021learning}. Hence, learning methods that are both trustworthy and powerful gradually arouse researchers' attention. This paper proposes a novel method that leverages the principles of physically constrained learning, aiming to infuse the model with a deeper understanding of the physical laws governing the system. Unlike conventional pure-learning approaches, which may overlook the conservation laws, our method utilizes distributed Port-Hamiltonian systems (dPHS) as a framework to incorporate these physical priors as inductive bias directly into the learning process.  

The Port-Hamiltonian framework is especially suited for modeling physical systems~\citep{nageshrao2015port}, as it provides a systematic representation of the energetic and dissipative characteristics intrinsic to the system under consideration. Further, Port-Hamiltonian systems have many beneficial properties such as being suitable for modeling multi-physics, inherently respecting energy conservation/dissipation, and being closed under composition.

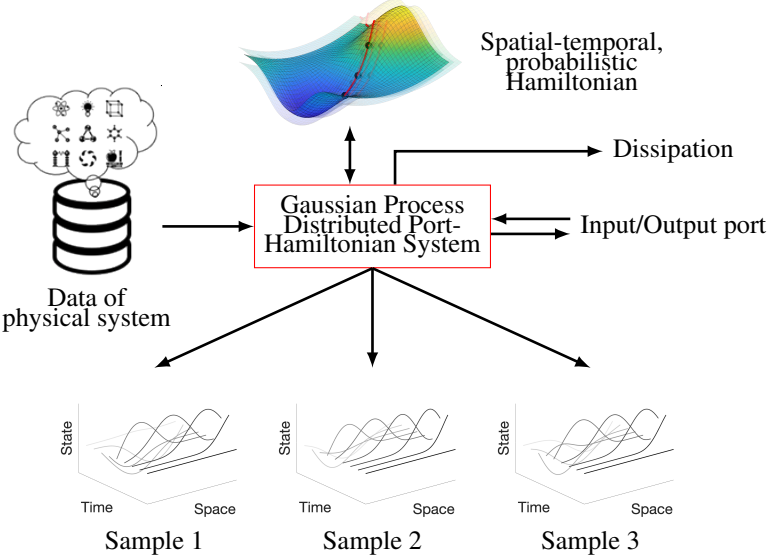
\begin{figure}[!b]
\begin{center}
	\input{figure/bsb_intro.tex}
	\caption{Gaussian Process Distributed Port-Hamiltonian (GP-dPHS) systems allow to learn a probabilistic spatial-temporal Hamiltonian based on collected data. The Port-Hamiltonian structure enables the inclusion of control inputs as well as dissipation. All sampled models of the GP-dPHS model are physically correct in terms of the evolution of the total energy in the system.}\label{fig:gpphs_intro}\vspace{-0.8cm}
\end{center}
\end{figure}

As an extension for PDEs, distributed Port-Hamiltonian systems have been introduced that enable representing a class of PDEs within the Port-Hamiltonian framework. The term distributed refers to the capability of these systems to operate across different variable domains, such as temporal and spatial. 
We capitalize on the data obtained from the observations of the PDE system. This data is then subsequently assimilated into a Gaussian Process, which aims to learn the Hamiltonian function of the system, thereby constructing a Bayesian data-driven distributed Port-Hamiltonian framework tailored for the PDE, see~\cref{fig:gpphs_intro}. Incorporating these physical constraints, our methodology not only conforms to established physical laws but also amplifies the model's capacity for generalization from sparse data, which enhances the desired learning efficiency. Furthermore, our proposed model incorporates mechanisms for uncertainty quantification. This feature is pivotal in ensuring the reliability of the model and can be exploited for safe model-based control approaches. In summary, we propose a physics-constrained learning approach that combines dPHS and Gaussian processes (GP-dPHS) to enable physically correct modeling for a class of PDEs with uncertainty quantification based on data.
\textbf{Related Work:} 
Prior research has explored the realm of learning the characteristics of Partial Differential Equation (PDE) systems, as evidenced by works such as \cite{raissi2019physics}, \cite{long2018pde} and \cite{chen2021solving}. However, these methodologies are predominantly learning-centric and tend to overlook the inherent physical properties of the systems in question. Reliably forecasting the evolution of the physics system requires uncertainty quantification due to the properties of physics, data, and the model itself. However, the approaches mentioned above ignore this point by offering merely predictions.
%%%%%%%%%%%
%Add here a sentence about PINNs and related approaches. Mention that they lack Uncertainty quantification.
%%%%%%%%%%%
Studies detailed in \cite{greydanus2019hamiltonian} and \cite{bertalan2019learning} have initiated the incorporation of the Hamiltonian framework to align the learning process with physical laws, a concept further extended to the domain of Port-Hamiltonian systems as demonstrated in \cite{desai2021port} and \cite{nageshrao2015port}. Despite these advancements, a notable deficiency within these methodologies is the absence of uncertainty quantification, a metric vital for assessing the learning's reliability. The Gaussian process (GP), distinguished by its Bayesian underpinnings, has been introduced to the learning of generating functions within Hamiltonian systems as delineated in \cite{karniadakis2021physics}, and has been applied to Port-Hamiltonian Systems in \cite{beckers2023data}. These approaches, however, are not able to accommodate the learning of PDE systems, signifying a gap in the current state-of-the-art that presents an opportunity for significant contributions to the field.
\\
\textbf{Contribution}:
We propose a physics-constrained learning approach, GP-dPHS, that combines dPHS and Gaussian processes. This approach is beneficial in multiple ways: i) it enables physically correct modeling for a class of PDEs in terms of the evolution of energy, ii) it is highly expressive due to the nonparametric nature of the GP and iii) it allows us to quantify the uncertainty of the model. The proposed model is in particular suitable for modeling low-dimensional but highly nonlinear PDEs and opens the door for the design of robust controllers based on the provided uncertainty of the proposed model. The detailed reasoning can be found in the following parts.\\ 

This paper details the theoretical underpinnings of our approach in \ref{sec:def}, elucidates the modeling process in \ref{sec:mod}, and presents empirical results in \ref{sec:eval} that showcase the efficacy of incorporating physical priors in learning nonlinear PDE systems.

\section{Background and Problem Formulation}\label{sec:def}
In this section, we briefly describe the class of distributed Port-Hamiltonian systems, revisit GP regression, and introduce the problem setting. 
%%%%%%%%%%%%%%%%%%%%%%%%%%%%%%%%%%%%%%%%%%%%%%%%%%
\subsection{Distributed Port-Hamiltonian Systems}\label{sec:dphs}
Composing Hamiltonian systems with input/output ports leads to a so-called Port-Hamiltonian system, which is a dynamical system with ports that specify the interactions of its components. In~\cite{macchelli2004port}, the classical finite-dimensional Port-Hamiltonian formulation of a dynamical system is generalized in order to include the distributed parameter and multivariable case. In contrast to finite-dimensional PHS, the interconnection, damping, and input/output matrices are replaced by matrix differential operators that are assumed to be constant, i.e., no explicit dependence on the state (energy) variables is considered. As in finite dimensions, given the Stokes–Dirac structure, the model of the system easily follows once the Hamiltonian function is specified. The resulting class of infinite-dimensional systems in Port-Hamiltonian form is quite general, thus allowing the interpretation of many classical PDEs within this framework and the description of several physical phenomena, such as the heat conduction, piezoelectricity, and elasticity. In the following, we recall the definition of dPHS from~\cite{macchelli2004port}.

Denote by $\Z$ a compact subset of $\R^n$ representing the
spatial domain of the distributed parameter system. Then,
denote by $\mathcal{U}$ and $\mathcal{V}$ two sets of smooth functions from $\Z$
to $\R^{n_u}$ and $\R^{n_v}$, respectively.
\\
\begin{defn}
A constant matrix differential operator of order $N$ is a map $L$ from $\mathcal{U}\to\mathcal{V}$ such that, given $u=(u^1,\ldots,u^{n_u})\in\mathcal{U}$ and $v=(v^1,\ldots,v^{n_v})\in\mathcal{V}$
\begin{align}
    v=Lu  \Longleftrightarrow v^b\coloneqq \sum_{\#\alpha=0}^N P_{a,b}^\alpha D^\alpha u^\alpha
\end{align}
where $\alpha\coloneqq\{\alpha_1,\ldots,\alpha_d\}$ is a multi-index of order $\#\alpha\coloneqq \sum_{i=1}^d \alpha_i$, $P^\alpha$ are a set of constant $n_u\times n_v$ matrices and $D^\alpha\coloneqq \partial_{z_1}^{\alpha_1}\cdots\partial_{z_d}^{\alpha_d}$ is an operator resulting from a
combination of spatial derivatives. Note that, in (6), the sum
is intended over all the possible multi-indices $\alpha$ with order $0$ to $N$ and implicitly on a from $1$ to $q$.
\end{defn}

\begin{defn}\label{defn:adjoint}
Denote by $J$ a constant matrix differential operator. Then, $J$ is skew-adjoint if and only if its formal adjoint $J^*$ is the map from $\mathcal{V}$ to $\mathcal{U}$ such that
\begin{align}
    u=-J^*v  \Longleftrightarrow u^b\coloneqq -\sum_{\#\alpha=0}^N (-1)^{\#\alpha}P_{a,b}^\alpha D^\alpha v^\alpha.
\end{align}
\end{defn}
\begin{defn}
Let $\delta_\x$ denotes the functional derivative operator such that
\begin{align}\label{derivative}
    \int \delta_x F(\z) \phi(\z) \; d\z
&= \lim_{\varepsilon\to 0}\frac{F[\x+\varepsilon \bm\phi]-F[\x]}{\varepsilon}= \left [ \frac{d}{d\varepsilon}F[\x+\varepsilon \bm\phi]\right ]_{\varepsilon=0},
\end{align}
where $\bm\phi$ is an arbitrary function.
\end{defn}
\begin{defn}
Let $\mathcal{X}$ be the space of
vector-valued smooth functions on $[0,\infty]\times \mathcal{Z}$ (energy
configurations), and $\mathcal{F}_d$ be the space of vector value smooth
functions on $\mathcal{Z}$ (distributed flows). Assume that $\mathcal{E}_d\equiv\mathcal{F}_d$
is its dual (distributed efforts). Moreover, denote by $J$ a skew-adjoint constant differential
operator, and by $G_d$ a constant differential operator.

Let $\mathcal{H}\colon\X\to\R$ be the
Hamiltonian functional such that
\begin{align}
    \mathcal{H}(\x)=\int_\mathcal{Z} H(z,x)dV,
\end{align}
where $H\colon\Z\times\X\to\R$ is the energy density. Denote by $\mathcal{W}$ the space of vector-valued smooth functions on $\partial\mathcal{Z}$ representing the boundary terms $\mathcal{W}\coloneqq \{w\vert w=B_\mathcal{Z}(\delta_\x \mathcal{H},\u)\}$ defined by the boundary operator $B_\mathcal{Z}$. Then, the general formulation of a multivariable
dPHS $\Sigma$ with constant
Stokes–Dirac structure is fully described by\footnote{Vectors~$\bm a$ and vector-valued functions~$\bm f(\cdot)$ are denoted with bold characters. Matrices are described with capital letters. $I_n$ is the $n$-dimensional identity matrix and $0_n$ the zero matrix. The expression~$A_{:,i}$ denotes the i-th column of $A$. For a positive semidefinite matrix $\Lambda$, $\|x - y\|_{\Lambda}^2 = (x - y)^\top \Lambda (x-y)$.  $\R_{>0}$ denotes the set of positive real number whereas $\R_{\geq 0}$ is the set of non-negative real numbers. $\C^i$ denotes the class of $i$-th times differentiable functions. The operator $\nabla_\x$ with $\x\in\R^n$ denotes $[\frac{\partial}{\partial x_1},\ldots,\frac{\partial}{\partial x_n}]^\top$}
\begin{align}\label{for:pch}
\Sigma(J,R,\mathcal{H},G)=\begin{cases}
    \dxdt=(J-R)\delta_\x \mathcal{H}+G_d\u\\
\y=G_d^* \delta_\x \mathcal{H}\\
w=B_\mathcal{Z}(\delta_\x \mathcal{H},\u),
\end{cases}
\end{align}
where $R$  is a constant differential operator taking into account energy dissipation and $(\u,\y)\in\mathcal{F}_d\times\mathcal{E}_d$. With the state $\x(t,\bm{z})\in\R^n$ (also called energy variable) at time $t\in\R_{\geq 0}$ location $\bm{z}\in\mathcal{Z}$ and the I/O ports $\u,\y\in\R^m$.
\end{defn}
\begin{rem}
We focus here on the multivariate distributed Port-Hamiltonian system with constant
Stokes–Dirac structure.
\end{rem}
 Generally speaking, the interconnection of the elements in the dPHS is defined by the $J$ matrix, whereas the Hamiltonian $H$ characterizes their dynamical behavior. In PDE systems,  the constitution of the $J$ matrix predominantly involves partial differential operators. The port variables~$\u$ and $\y$ are conjugate variables in the sense that their duality product defines the energy flows exchanged with the environment of the system, for instance, currents and voltages in electrical circuits or forces and velocities in mechanical systems, see~\cite{van2000l2} for more information.

\subsection{Gaussian Process Regression}\label{sec:GPIntro}

The Gaussian Process model is a nonparametric model widely used in machine learning to perform regression and prediction tasks, see \cite{williams1995gaussian}. 

Compared to other learning methods in regression tasks, GPs enable a natural quantification of uncertainty instead of just point estimates. The observations of the system are usually limited in practice. However, GPs can provide good predictions with relatively small data sets by leveraging the structure of the data through the kernel. 

A GP can be fully defined by giving a mean function $m$ and a covariance function $k$. For each pair of observation $\x,\x'$, the kernel function $k(\x,\x)$ describes the covariance between these points. Before observing any data, we define a prior distribution over functions. For a set of input points, the prior is a multivariate (zero-mean) Gaussian process: $f\sim GP(m(\x),k(\x,\x'))$. Let $\{X = \x^{\{1\}}, \x^{\{2\}}, ..., \x^{\{N\}}\}$ denotes the input data set, together with the output data set as$\{Y = \y^{\{1\}}, \y^{\{2\}}, ..., \y^{\{N\}}\} $, to compose the training data $\mathcal{D}$ with $N$ data points. The output data might be corrupted by zero-mean Gaussian noise with variance $\sigma_{n}^2$. After obtaining the data $\mathcal{D}$, we update our belief about the function $f$ using the Bayes' Theorem to get the posterior distribution given a test input $x^*$. The posterior is again a Gaussian process with updated mean and covariance
\begin{align}
    \mu(y^*|\x^*) = k(\x^{*},X)[K(X,X)+\sigma_{n}^2I]^{-1} Y   \label{for:PostMean}
\end{align}\vspace{-0.8cm}
\begin{align}
    \Sigma(y^*|\x^*) &= k(\x^*,\x^*) - K(\x^*,X)[K(X,X)+\sigma_{n}^2I]^{-1}K(\x^*,X). \label{for:PostVar}
\end{align}
In the formulation, $K(\x^*,X)$ is the covariance between the new point and the training points, and $k(\x^{*},\x^*)$ is the prior variance at the new point.
The predictive distribution $p(y^*|\x^*, X, Y)$ provides both a mean estimate and a measure of uncertainty for the predictions at a new point $\x^*$, which is the essence of uncertainty quantification in the GP regression learning model.

\subsection{Problem Setting}\label{sec:ps}
We are considering building a physical-constrained learning system for a class of PDE systems. For many physical systems, their dynamics can be written in distributed Port-Hamiltonian form, see \cref{for:pch}. In our case, the evolution of the states of a PDE system $\x(t,z)\in\R^n$ over time $t\in\R^n_{\geq0}$ and spatial domain $z\in\Z$ should be defined by the equation
with initial state $\x(0,z)$ as described in~\cref{sec:dphs}.
The Hamiltonian functional $\mathcal{H}\in \C^{\infty}$ is assumed to be \textit{completely unknown} due to unstructured
uncertainties in the system imposed by, e.g., nonlinear stress/strain curve in soft materials, physical coupling effects, or highly nonlinear electrical and magnetic fields. Thus, we aim to learn a dPHS model
\begin{align}
	\dxdt&=(J-R)\delta_\x \hat{\mathcal{H}}+G_d\u
\end{align}
with an estimated Hamiltonian functional $\hat{\mathcal{H}}$ based on observations of the system~\cref{for:pch}.
To address the problem, we need to introduce the following assumptions.
\begin{assum}\label{assum:1}
The structure of the system matrices $J$, $R$, and $G$ are assumed to be known, with the exception of a finite set of parameters $\Theta \subset \mathbb{R}^{n_\Theta}$, where $n_\Theta$ is the total number of unknown parameters inside of these matrices.
\end{assum}
\begin{assum}\label{assum:2}
We can observe the state of the PDE system \cref{for:pch} at some temporal $t_i$ and spatial points $\z_j$ to generate a set of observations $\{\x(t_1,\z_1),\ldots,x(t_1,\z_{N_\z})\,\\\ldots,\x(t_{N_t},\z_1), \ldots,\x(t_{N_t},\z_{N_z})\}$.
\end{assum}
\begin{assum}\label{assum:3}
There exists a unique and smooth solution $\x(t,z)$ for the PDE system under the known boundary conditions $w$~\cref{for:pch}.
\end{assum}
\Cref{assum:1} is only mildly restrictive since $J$ and $G$ are typically matrix differential operators, given by prior knowledge of the PDE system to be learned. Even though we imply knowing the dissipation matrix $R$, we allow unknown parameters such that only the general structure needs to be known, e.g., the friction model but not the parameters. \Cref{assum:2} ensures that we can collect data from the PDE system, requiring observability of the state. If that is not the case, an observer needs to be implemented. Finally, \cref{assum:3} guarantees the existence of a solution so that the derivation of a model is meaningful.

%%%%%%%%%%%%%%%%%%%%%%%%%%%%%%%%%%%%%%%%%%%%%%%%%%
\section{Gaussian Process dPHS}\label{sec:mod}

In this section, we propose the Gaussian Process Distributed Port-Hamiltonian system (GP-dPHS). The objective of the proposed method is to develop a data-driven model for the learning of low-dimensional but nonlinear PDE systems, which is characterized as a distributed Port-Hamiltonian system with a structure as delineated in \cref{for:pch}. The main challenge of extending this approach to high-dimensional problems is that the later represented kernel function leads to a Gram matrix $\bm{K}$, whose size grows with an increase of the dimension of system state $\x$. Thus, calculating the inverse of the covariance matrix would be relatively expensive. First, we create a dataset that consists of state and state derivatives of the PDE dynamics. On the basis of this data set, we can train a GP model with a physics-constrained kernel function to learn the PDE dynamics. Finally, we use the model to predict future states.

\subsection{GP-dPHS data generation}
As stated in the problem setting, we consider the observations as follows
\begin{align}
    \mathcal{D} = \{t_i,\z_j,\x(t_i,\z_j),\u(t_i)\}_{i=1,j=1}^{i=N_t,j=N_z},
\end{align}
where $\x(t_i,\z_j)$ encapsulates the observed state at time $t_i$ and spatial point $\z_j$ for $N_t$ time steps and $N_z$ spatial points, corresponding to the input sequence $\{\u(t_1),\\\u(t_2), ..., \u(t_{N_t})\}$. With this time-sequenced observational data $\mathcal{D}$ at our disposal, the first step in modeling the PDE system through the GP-dPHS is the procurement of the system's state derivatives $\frac{\partial \x}{\partial t}$. 
Furthermore, we typically have only a small set of spatial observation points $N_z$ due to technical limitations. Referencing the equation detailed in \cref{derivative}, and employing a smooth kernel $k$ such as the squared exponential kernel, our approach is to construct $n$ distinct GPs for the training sets $\mathcal{D}$. To expound for each $l$-th dimension of the state vector, we have a training set 
$\mathcal{D}_l = \{t_i,\z_j,x_l(t_i,\z_j),\u(t_i)\}_{i=1,j=1}^{i=N_t,j=N_z},$ where $l=1,\ldots,n$. Based on these data sets, we train $n$ separate GP models with $\{t_i,\z_j\}_{i=1,j=1}^{i=N_t,j=N_e}$ as input data and $\{x_l(t_i,\z_j)\}_{i=1,j=1}^{i=N_t,j=N_z}$ as output data. 

\begin{figure}[H]
  \centering
    \includegraphics[width=0.45\linewidth]{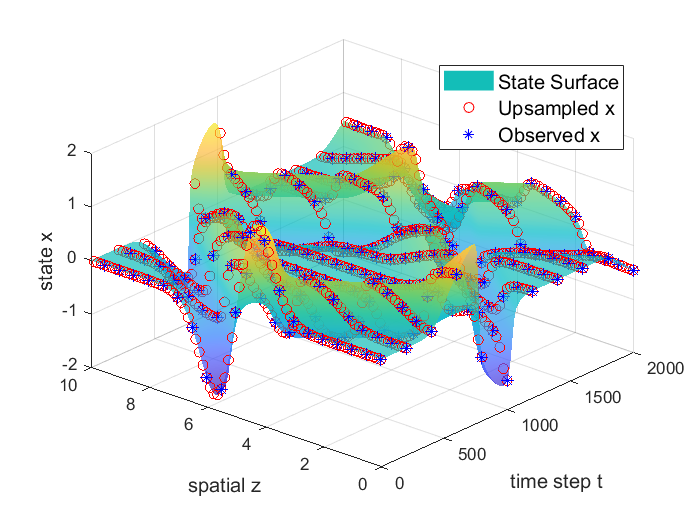}\vspace{-0.1cm}
  \caption{Example of states observations and the upsampling process.}
  \label{fig:observation}\vspace{-0.5cm}
\end{figure}

By differentiating the kernel function $k$, we can then acquire the desired derivative $\frac{\partial x_l}{\partial t}$ of the corresponding states $x_l$. The properties of $\frac{\partial x_l}{\partial t}$ can be fully described by the GP mean function $\mu(\frac{\partial x_l}{\partial t}|t,\mathcal{D}_l)$ and the variance function $\Sigma(\frac{\partial x_l}{\partial t}|t,\mathcal{D}_l)$. Thus, we can obtain a new dataset $\mathcal{E}$, which contains the states $X=[\tilde{\x}(t_1),\ldots,\\\tilde{\x}(t_{N_t})]$ and the state derivative $\dot{X}=[\frac{\partial \tilde{\x}(t_1)}{\partial t},\ldots,\frac{\partial \tilde{\x}(t_{N_t})}{\partial t}]$, where $\tilde{\x}(t_i)=[\x(t_i,\\\z_1)^\top,\ldots,\x(t_i,\z_{N_e})^\top]^\top$ denotes the spatially stacked state at time $t_i$. Note that we upsampled the number of spatial points from $N_z$ to $N_{e}>>N_z$ by evaluating the GP models on new spatial points. We visualize this step in~\cref{fig:observation}.

\subsection{Modeling}
After the generation of the data, we now introduce the GP-dPHS model.
The central concept underpinning this approach involves the employment of a Gaussian process to approximate the unknown Hamiltonian functional while simultaneously treating the parametric indeterminacies within the matrices $J$, $R$, and $G$ as hyperparameters. This requirement is accommodated by exploiting the characteristic that GPs are invariant under linear transformations \citep{jidling2017linearly}, which allows inclusion of the Hamiltonian functional derivative within the GP formulation. The dynamical behavior of the system is thus expressed as a linear transformation of the Hamiltonian within the Port-Hamiltonian (PH) model. This integration of the PHS dynamics into the GP is mathematically represented as
\begin{align}
\frac{\partial \x}{\partial t}\sim\mathcal{GP}(\hat{G}_\Theta \u,k_{dphs}(\x,\x')),\label{gp:frac}
\end{align}
where the system dynamics are encapsulated within the probabilistic model. Specifically, the new kernel function $k_{dphs}$ can be formally described as 
\begin{align}
    k_{dphs}(\x,\x') = \sigma^2_f(\hat{J}_\Theta-\hat{R}_\Theta)\delta_\x\exp\left(-\frac{||\x-\x'||^2}{2\varphi_l^2}\right)\delta_{\x'}^\top(\hat{J}_\Theta-\hat{R}_\Theta)^\top.
\end{align}
This kernel function is based on the squared exponential function as the Hamiltonian functional is a smooth function. The model can then be trained as a standard GP with the dataset $\mathcal{E}$~\cref{gp:frac} from the previous section. The matrices $J$, $R$ and $G$ of the dPHS system are inferred through the estimate $\hat{J}_\Theta,\hat{R}_\Theta,\hat{G}_\Theta$, where the unknown parameters given by $\Theta$ are treated as hyperparameters as introduced in \cref{sec:ps}. The unknown hyperparameters $\Theta,\varphi_l,\sigma_f$ are minimized by means of
the negative log marginal likelihood. In this way, we complete the training of the proposed GP-dPHS.
By exploiting again the invariance under linear transformation, we can rewrite~\cref{gp:frac} as 
\begin{align}\label{dPHSform}
\frac{\partial}{\partial t}
\x(t,\z)
=
(\hat{J}_\Theta-\hat{R}_\Theta)
\delta_\x \hat{\mathcal{H}}+\hat{G}_\Theta\u,
\end{align}
where the Hamiltonian $\hat{\mathcal{H}}$ follows a GP. It can be seen that the dynamics describes a distribution over distributed Port-Hamiltonian systems.
Analogously to the approach delineated in \cref{for:pch}, our strategy uses data-driven techniques to infer the characteristics of the distributed physical system while rigorously maintaining compliance with intrinsic physical constraints. This multifaceted learning is integrated to construct our Gaussian Process distributed Port-Hamiltonian systems, which embodies a comprehensive model encapsulating the dynamical interplay within diverse domains. 

\subsection{Prediction}
In pursuit of generating samples from the posterior distribution, we employ the joint distribution with mean-adjusted output data at test states $\x^*\in\R^n$
\begin{equation}
\begin{bmatrix}
X \\
\bm{f}(x^*) \\
\end{bmatrix}
=
\mathcal{N}
\left(
\hat{G}_\Theta\u,
\begin{bmatrix}
K_{dphs}(X,X) & K_{dphs}(X,\x^*)\\
K_{dphs}(X,\x^*)^T & K_{dphs}(\x^*,\x^*) \\
\end{bmatrix}
\right),
\end{equation}
which facilitates the procurement of a vector field $\bm{f}$ that represents the right-hand side of~\cref{dPHSform}. From this distribution, we could sample a $\bm{f}$, i.e., a sample of the learned dPHS system, and solve the resulting PDE with a numerical solver. However, the demands of numerical integration necessitate the ability to access this vector field at various points, which is computationally expensive~\citep{9632353}. Instead, the approach advocates to sample the estimated Hamiltonian functional variable named $\mathcal{\Hat{H}}$, which leads to~\cref{dPHSform}. Thus, we sample a deterministic Hamiltonian $\mathcal{\hat{H}}$ from the GP distribution and use a numerical PDE solver to compute a solution for~\cref{dPHSform}. Since a sample from GP is the deterministic function and the dPHS is designed to follow \ref{for:pch}, it is fully compliant with the Stokes-Dirac structure.  This procedure guarantees that the solution of the learned dynamics is physically correct in terms of the energy evolution, since it has to satisfy the dPHS form.
\section{Numerical evaluation}\label{sec:eval}

For evaluating the proposed physics-constrained learning methodology, we aim to learn the dynamics of the lateral vibration of a string with a nonlinear stress-strain curve. This system is succinctly represented by the ensuing general equation
\begin{equation}
    \frac{\partial^2 x}{\partial t^2}-s\left(\frac{\partial x}{\partial z}\right)\frac{\partial^2 x}{\partial z^2} =0,
\end{equation}
where $s$ denotes the stress/strain curve that is, for soft materials, a highly nonlinear function of the stress $\frac{\partial x}{\partial z}$. In our example, we assume that $s$ follows a sigmoid function. To use our proposed GP-dPHS model, we need to translate the system into dPHS form, where the functional Hamiltonian $\mathcal{H}$ is assumed to be unknown. Recalling equation \cref{for:pch}, it is postulated that the constant matrix differential operators $\mathcal{J}$, $\mathcal{R}$, and $\mathcal{G}$ are structurally known differential operators. In our setting, we are examining a string system with both ends fixed and the spatial domain $z\in[0,10]$. For the sake of simplicity, we assume no input and imply a nontrivial initial condition of the system to generate a meaningful dataset. We set the damping factor to $0.01$. Then, the desired system can be modeled via a dPHS in the following form
\begin{align}\label{eq:grtr}
     {\frac{\partial}{\partial t}}
     \begin{bmatrix}
        p(t,z) \\
        q(t,z) \\
    \end{bmatrix}
     =
    \underbrace{ \begin{bmatrix}
        - 0.01 &{\frac{\partial}{\partial z}} \\
        {\frac{\partial}{\partial z}}& 0 \\
    \end{bmatrix}}_{J-R}
    \delta_\x \mathcal{H},
\end{align}
where $p=\frac{\partial x}{\partial t}$ and $q=\frac{\partial x}{\partial z}$. This transformation is necessary to ensure that $J$ is an adjoint operator as defined in~\cref{defn:adjoint}. The Hamiltonian is given by $\mathcal H=\int_\mathcal{Z}\int s(q)q(t,z)dq+p(t,z)^2dz$. We refer to~\cref{eq:grtr} as the ground truth.

\begin{figure}[H]
  \centering
    \includegraphics[width=0.45\linewidth]{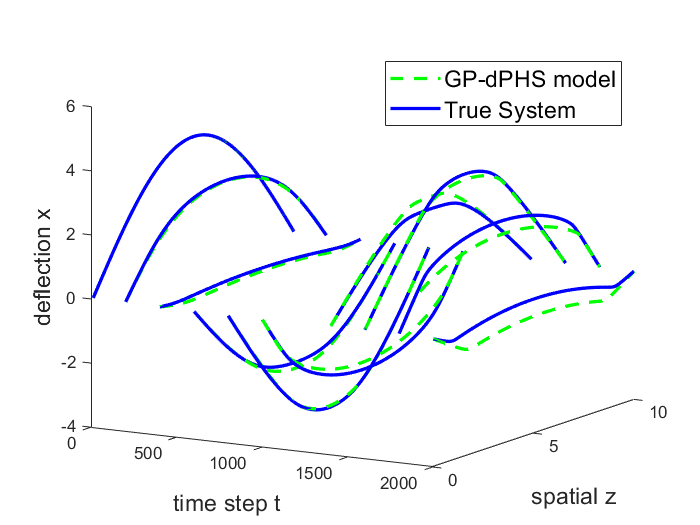}\vspace{-0.1cm}
    \includegraphics[width=0.45\linewidth]{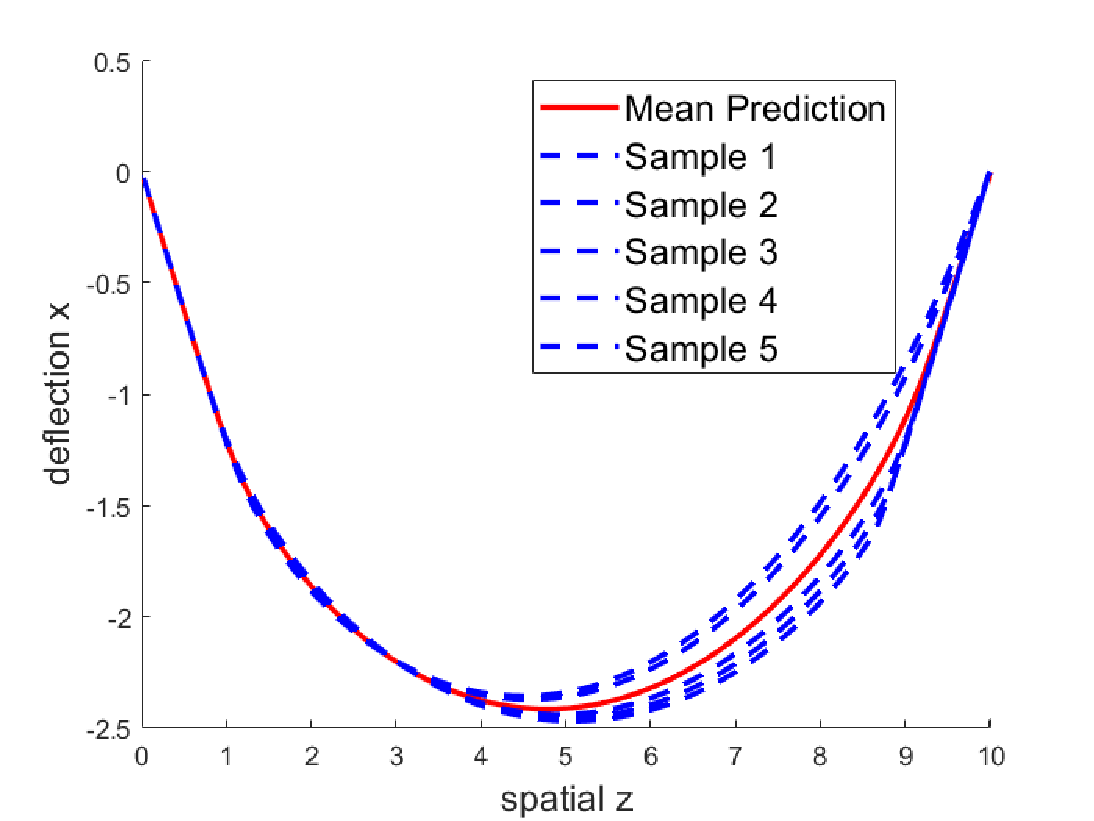}\vspace{-0.1cm}
  \caption{Graphical illustration of string PDE system. Observe on the left figure of each time slice; the deflection learned by GP-dPHS (cyan dotted) is close to the ground-truth trajectory on each time step t. The mean prediction of time slice $t = 600$ and five samples of the GP-dPHS model are shown on the right.}
  \label{fig:landscape}\vspace{-0.5cm}
\end{figure}

\begin{figure}[H]
  \centering 
  \includegraphics[width=0.45\linewidth]{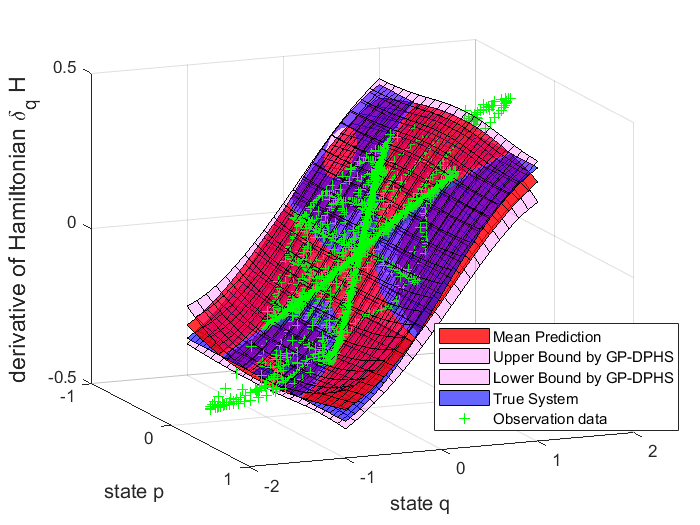}\vspace{-0.1cm}
  \includegraphics[width=0.45\linewidth]{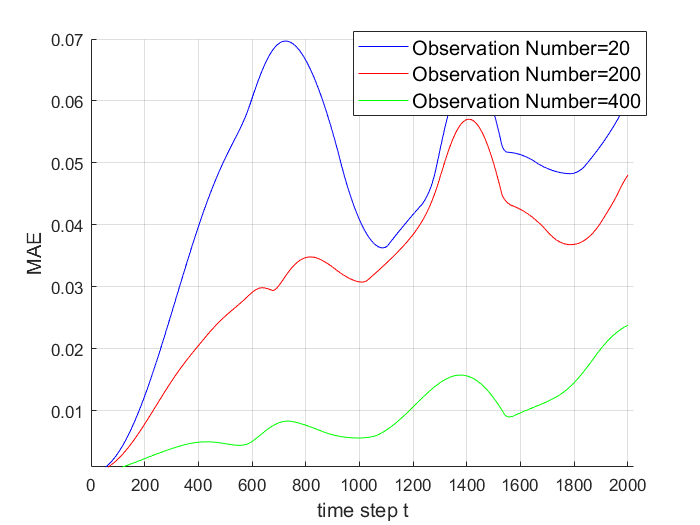}\vspace{-0.1cm}
  \caption{The left figure visualizes the nonlinear Hamiltonian derivative $\delta_q H$. The learned surface (red) is very close to the true surface (blue). The pink surfaces visualize the uncertainty of the Hamiltonian derivative under a 95\% prediction uncertainty guarantee. The figure on the right shows the model error for different numbers of training points}  \vspace{-5mm}\label{fig:H}
\end{figure}

Following the problem statement, we assume that the structure of~\cref{eq:grtr} is known, but the Hamiltonian is unknown to us. This example demonstrates that this is a reasonable assumption as the nonlinearity of the system is encapsulated in the Hamiltonian.  At the time point $t_0$, we set the initial condition to $x_0=\exp(-(z-5)^2)$. We observe the ground truth system for $20$ seconds with a time step of $\Delta t=0.01$ seconds. The spatial domain is observed at $N_z=$ 8 points. We upsample it to $N_e =$ 400 points by evaluating the GP models on new spatial points. To accelerate the training, we flatten all the observations, apply a downsample rate = 50 to all the data, and train the proposed GP-dPHS model using the steps mentioned above. Then, we compare the GP-dPHS model with the ground truth for a new initial condition, i.e., $x_0=\sin(\frac{10z}{\pi})$. The movements of the string are visualized in~\cref{fig:landscape}. It can be seen that the mean prediction of our proposed GP-dPHS model accurately mimics the ground truth.~\Cref{fig:H} shows the nonlinear ground truth and the learned derivative of the Hamiltonian $\mathcal{H}$. As shown in the figure, our proposed method learned the desired Hamiltonian surface, which is very close to the ground truth. Also, we visualize 95\% confidence interval guarantees of uncertainty in the same figure to show that the proposed method offers confidence intervals for uncertainty quantification instead of giving a deterministic prediction that can be inaccurate, 

As a physics-constrained method, we expect that the model generalizes well even for small datasets. The following figure on the right side shows that, using very limited observations, our proposed method is capable of learning a relatively accurate physics-constrained model. Note, that independently of the number of data points, the GP-dPHS model always leads to physically correct solutions.
\vspace{-5mm}
\section{Conclusion}
In this paper, we propose GP-dPHS, a physics-constrained data-driven method for learning a class of PDE systems based on data. Our approach guarantees the physical correctness of the model, generalizes well for small datasets, and provides uncertainty quantification. In a numerical example, we demonstrate the proposed approach for a simple nonlinear PDE problem. In the future, we will take advantage of the dPHS framework to design a physics-based controller for PDE systems.

\bibliographystyle{IEEEtran}
\bibliography{mybib.bib}

\end{document}

%% file: figure/bsb_intro.tex
\tikzsetnextfilename{bsb_intro}
\begin{tikzpicture}[auto, node distance=2cm,>=latex]
	\tikzstyle{block} = [draw, fill=white, rectangle,  line width=1pt, 
    minimum height=1.5em, minimum width=2em, font=\small,align=center, inner sep=3pt]
	\tikzstyle{input} = [coordinate]

    % We start by placing the blocks
    \node [transform canvas={yshift=0.6cm},inner sep=0pt, node distance=3cm,label={[label distance=-0.6cm,font=\small,align=center]-90:Data of\\[-1ex]physical system}] (data) {\includegraphics[width=2cm]{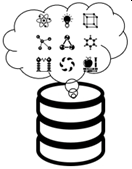}};
    
    \node [draw=red, right of=data, node distance=3.8cm,align=center,font=\small] (energy) {Gaussian Process\\[-1ex]Distributed Port-\\[-1ex]Hamiltonian System};
    \draw [->,line width=1pt]  (data.east) --  (energy.west);
    \node[inner sep=0pt, above of=energy, node distance=2.2cm,xshift=-0.3cm,label={[label distance=0cm,font=\small,align=center,text width=6.8em]0:Spatial-temporal,\\[-1ex]probabilistic\\[-1ex]Hamiltonian}] (man) {\includegraphics[width=3cm]{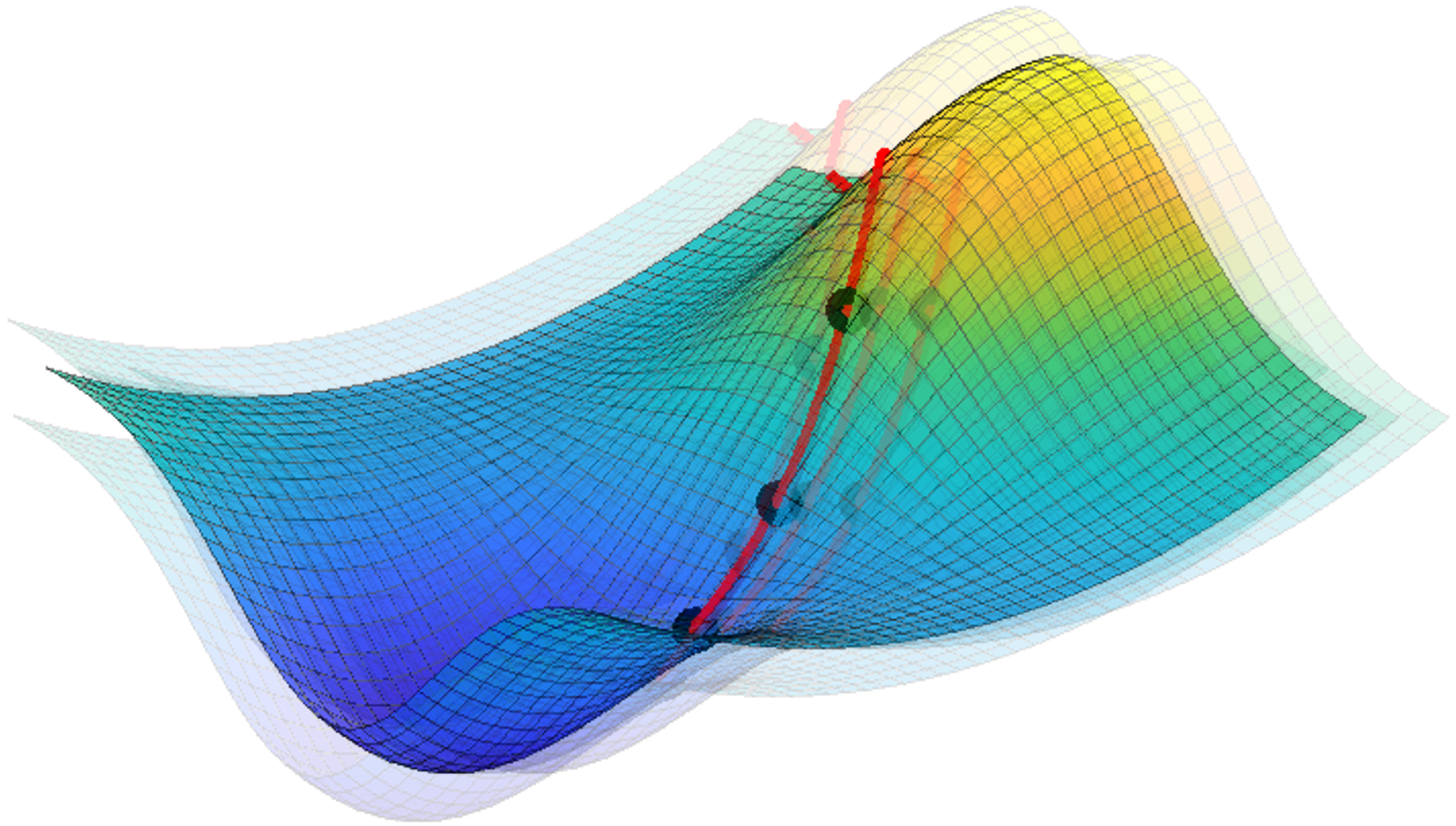}};
    \node[inner sep=0pt, below of=energy, node distance=2.9cm,label={[label distance=0cm,font=\small,align=center]-90:Sample 2}] (reali) {\includegraphics[width=2.7cm]{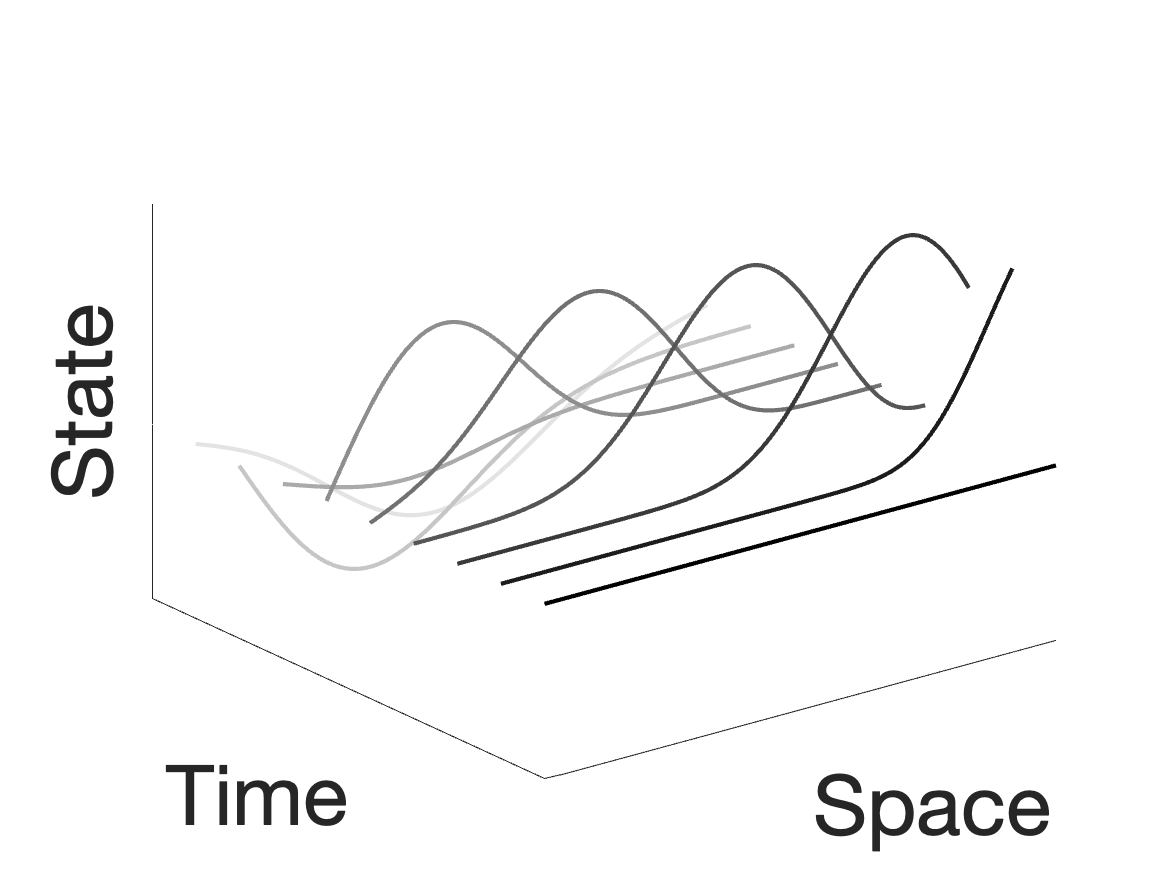}};
    \node[inner sep=0pt, left of=reali, node distance=2.9cm,label={[label distance=0cm,font=\small,align=center]-90:Sample 1}] (reali1) {\includegraphics[width=2.7cm]{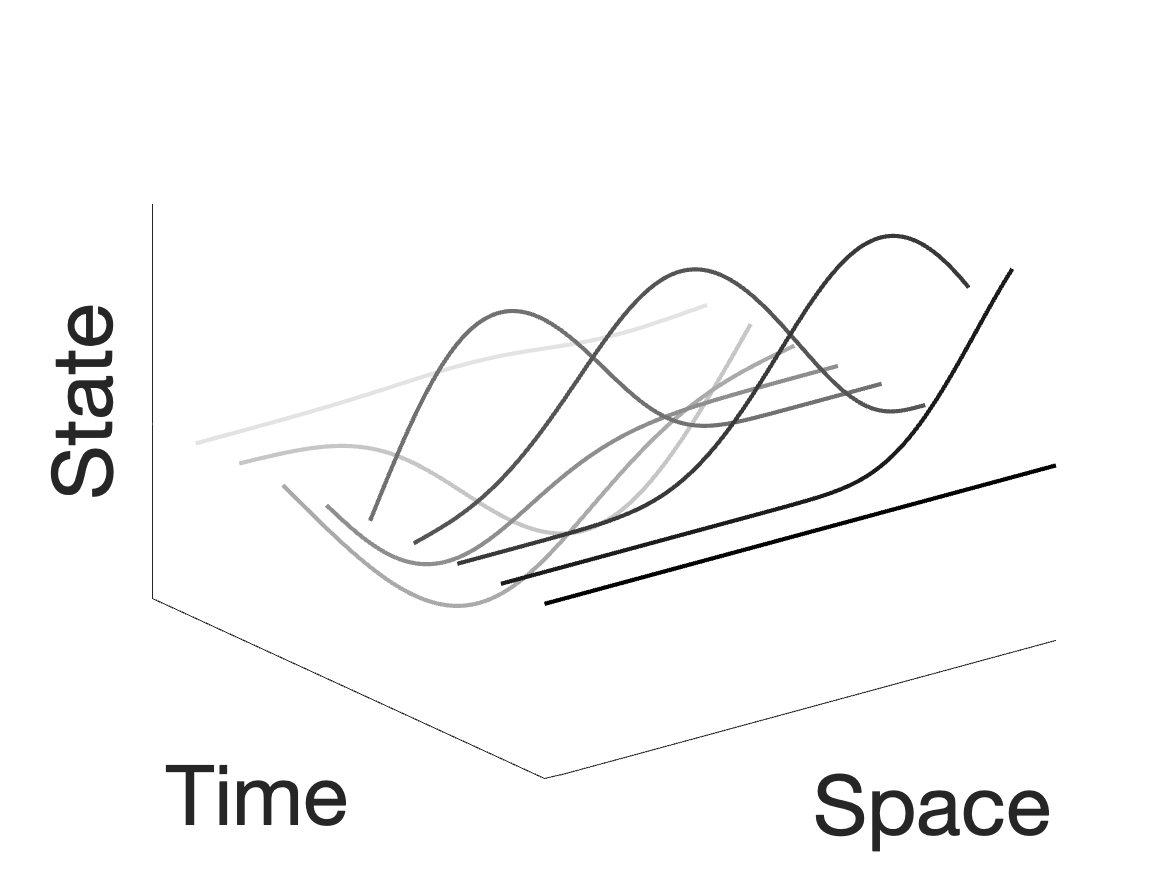}};
    \node[inner sep=0pt, right of=reali, node distance=2.9cm,label={[label distance=0cm,font=\small,align=center]-90:Sample 3}] (reali2) {\includegraphics[width=2.7cm]{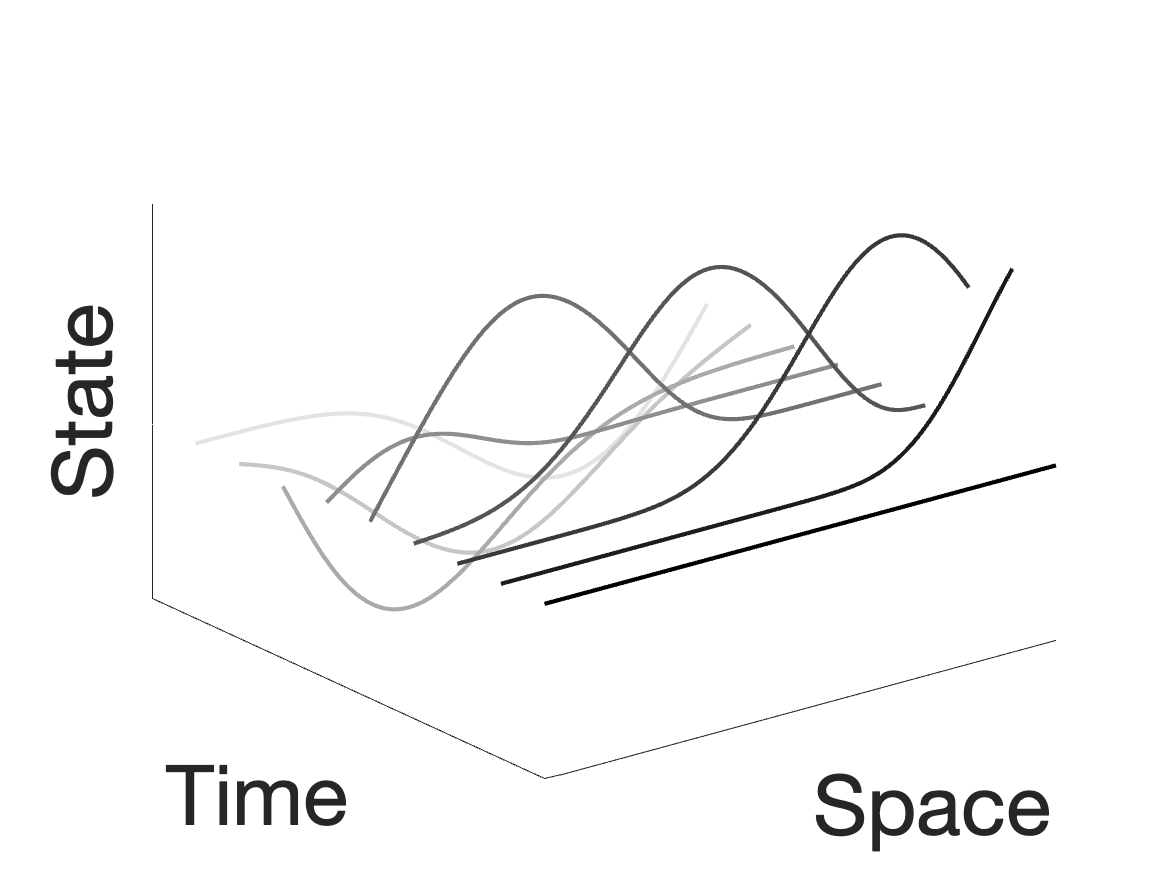}};
    \node [draw=none, right of=energy, node distance=4cm,align=center,font=\small] (input) {Input/Output port};
    \node [draw=none, above of=input, node distance=1cm,align=center,font=\small] (diss) {Dissipation};
    \draw [->,line width=1pt]  (energy.south) --  (reali.north);
    \draw [->,line width=1pt]  (energy.south) --  (reali1.north);
    \draw [->,line width=1pt]  (energy.south) --  (reali2.north);
    \draw [<->,line width=1pt] (man.south) --  ([xshift=-0.3cm]energy.north);
    \draw [<-,line width=1pt] ([yshift=-0.1cm]input.west) --  ([yshift=-0.1cm]energy.east);
    \draw [->,line width=1pt] ([yshift=0.1cm]input.west) --  ([yshift=0.1cm]energy.east);
    \draw [->,line width=1pt]  ([xshift=0.3cm]energy.north) |-  (diss.west);
\end{tikzpicture}